%% file: paper.tex
\documentclass[a4paper]{article}
\usepackage{arxiv}
\usepackage[labelfont=bf]{caption}
\usepackage[utf8]{inputenc}
\usepackage{makeidx}
\usepackage{units}
\usepackage{enumitem}
\usepackage{textcomp}
\usepackage{amsmath}
\usepackage{bm}
\usepackage{amssymb}
\usepackage{stackrel}
\usepackage{graphicx}
\usepackage{wasysym}
\usepackage{esint}
\usepackage{cite}
\usepackage{etoolbox} 

\usepackage[font=small,labelfont=bf]{caption}
\usepackage[unicode=true,
 bookmarks=false,
 breaklinks=false,pdfborder={0 0 1},backref=section,colorlinks=false]
 {hyperref}
\makeatletter

\usepackage{makeidx}
\makeatother

\begin{document}
\input{titleauthors.tex}

\input{abstract.tex}

\input{introduction.tex}

\input{stateoftheart.tex}

\input{methodology.tex}

\input{experiments.tex}

\input{conclusion.tex}

\input{bibliography.tex}
\end{document}

%% file: titleauthors.tex
\institute{Institute for Flight Mechanics and Controls, University of Stuttgart, Germany \email{\{eric.price,aamir.ahmad\}@ifr.uni-stuttgart.de}\and Max Planck Institute for Intelligent Systems, Tübingen, Germany.}

\docdate{2023-02-19}

\author{Eric Price\inst{1,2} \and Aamir Ahmad\inst{1,2}}

\title{Accelerated Video Annotation driven by Deep Detector and Tracker}\maketitle

%% file: abstract.tex
\begin{abstract}Annotating object ground truth in videos is vital for several 
downstream tasks in robot perception and machine learning, such as 
for evaluating the performance of an object tracker or training an image-based 
object detector.
The accuracy of the annotated instances of the moving objects on every image
frame in a video is crucially important. Achieving that through manual 
annotations is not only very time consuming and labor intensive, but is also 
prone to high error rate. State-of-the-art annotation methods depend on 
manually initializing the object bounding boxes only in the first frame and 
then use classical tracking methods, e.g., adaboost, or kernelized correlation 
filters, to keep track of those bounding boxes. These can quickly drift, thereby 
requiring tedious manual supervision. In this paper, we propose a new 
annotation method which leverages a combination of a learning-based detector 
(SSD) and a learning-based tracker (RE$^3$). Through this, we significantly 
reduce annotation drifts, and, consequently, the required manual supervision. 
We validate our approach through annotation experiments using our proposed 
annotation method and existing baselines on a set of drone video frames. Source code 
and detailed information on how to run the annotation program can be found at
\url{https://github.com/robot-perception-group/smarter-labelme}\end{abstract}

%% file: introduction.tex
\section{Introduction}

\begin{figure}
\begin{centering}
\includegraphics[width=1.0\columnwidth]{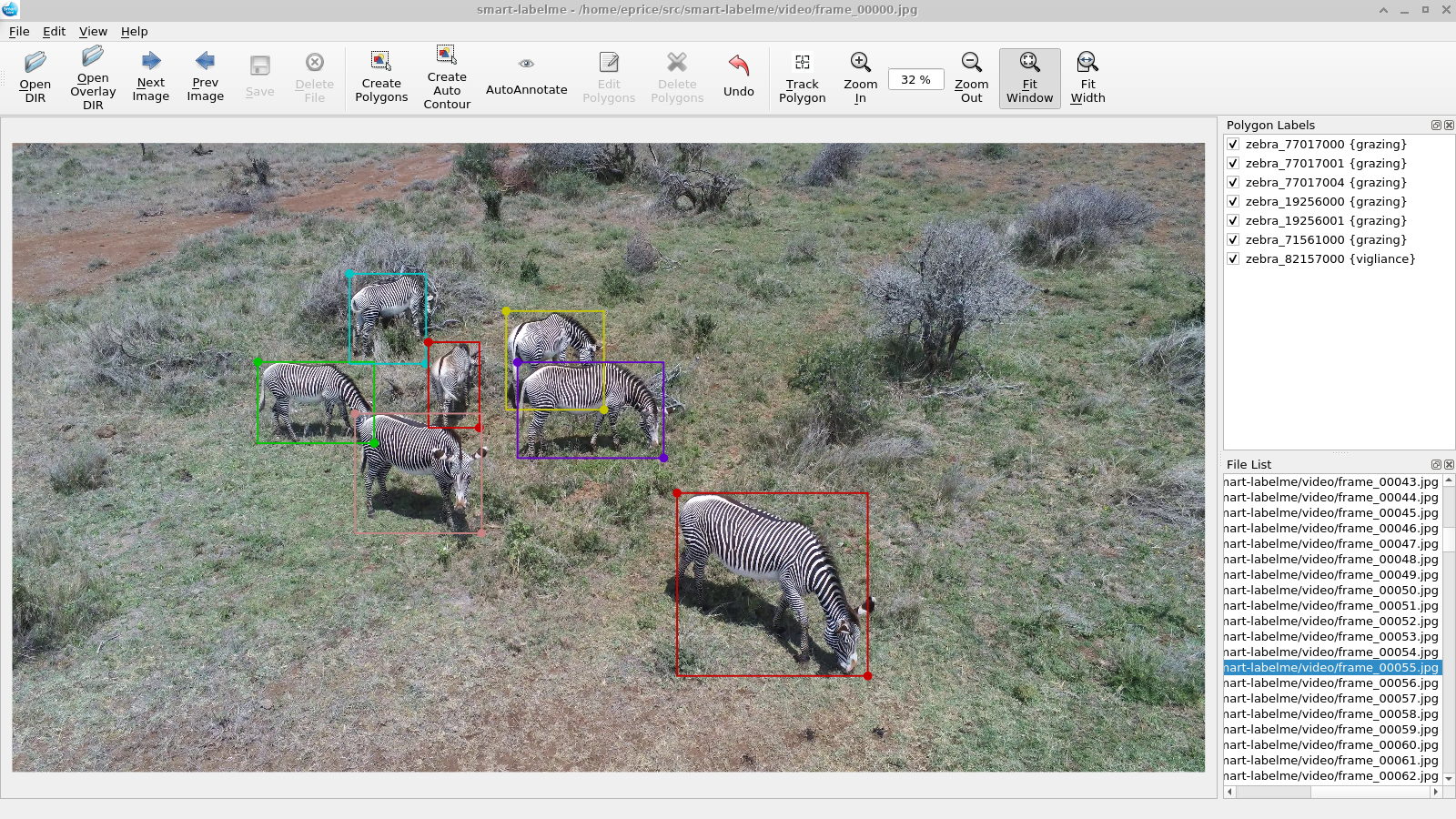}
\par\end{centering}
\caption{\label{fig:Titlefigure}Screenshot of \emph{Smarter-labelme} using SSD-Multibox 
to auto annotate zebras.}
\end{figure}

To annotate ground-truth of objects in a video, the annotator must solve two 
tasks. First, detection of new relevant object instances in each video frame. 
Second, tracking of already known object instances in subsequent video 
frames. A graphical user interface (GUI) helps human annotators with 
these tasks. It can display all relevant information and allow the user to add 
and modify annotations as quickly as possible. Using visual detection to pre-annotate data -
either on the fly or in a preprocessing step - is also common  \cite{8748669,4515873}. State-of-the-art open-source programs, like 
\emph{Smart-labelme} \cite{smart-labelme2020}, are particularly promising for the 
purpose of video annotations, since they implements tracking of already 
annotated objects across frames using classical tracking methods, e.g., 
adaboost, or kernelized correlation filters. Such trackers often suffer from 
``drift''. What the tracker considers as a part of the object or not can change 
in every frame, e.g., a tracker initialized to track a person might drift to 
eventually track only his/her head or a similar visually easy-to-distinguish 
part.

In this paper, we present a new annotation method, \emph{Smarter-labelme}, by
combining a deep learning-based tracker with a deep learning-based detector
that compensates for this drift. While, such techniques have been used for
visual tracking methods, to the best of our knowledge they have not been
introduced for a freely-available video annotation tool. Furthermore, the
visual detector in our method also address the challenge of detecting new and
relevant objects to begin with. In our approach, the initial detection is
aided by applying SSD-Multibox detector, pre-trained on the MSCOCO dataset on a
diverse set of common object classes. The detector is applied in a hierarchical
overlapping sliding window technique with global, exclusive, non-maximum
suppression to avoid re-detecting already-annotated object instances, truncated
objects or multiple detections per object. Instances are assigned a globally
unique ID consisting of the detected class name and unique random numeric
identifier. For tracking, we use Re$^3$, to pre-calculate probable bounding box
annotations.  SSD-Multibox detector is then used to correct this bounding box
for drift if a detection with sufficiently high confidence and Jaccard overlap
is found.  Annotations are stored in per-video-frame JSON files with consistent
instance labels over time. This allows post-processing of annotations,
including fine-tuning of the networks that use the annotated data.

%% file: stateoftheart.tex
\section{Related Works}

Techniques and tools for video annotation have been extensively studied and
applied \cite{8748669}. Several approaches use cloud-based computation to
crowd-source annotation. However, internet bandwidth considerations limit the
usability on video. Recent works on video annotation
\cite{Kuznetsova_2021_WACV} exploit computer vision aided interpolation between
human annotated key frames. In this work, we extend \emph{labelme}
\cite{labelme2016}, an open source video annotation tool which has recently
been improved \cite{smart-labelme2020} to automatically track objects using a
traditional computer vision technique provided by OpenCV \cite{6240859,4515873}
and is now called \emph{Smart-labelme}. We build on this tool with state-of-the-art
methods for both object detection and tracking. Visual object tracking in
video has also been extensively studied \cite{10.1145/3309665,9913708}. Our choice
is a light weight network, Re$^3$ \cite{8253805}, that uses a recurrent siamese
approach. Deep convolutional networks have also revolutionized visual object
detection \cite{10028728, WU202039}. For bounding box detection, we use
SSD-Multibox \cite{10.1007/978-3-319-46448-0_2}, since it has a good tradeoff
between speed and accuracy. This network is no longer state-of-the-art mainly
due to bad performance on very small objects, but we can circumvent this
shortcoming by optimal scaling of detection regions for object tracking,
inspired by \cite{8394622} and within a multi-scale sliding window object
detection approach, inspired by \cite{8280135}. However, instead of
parallelizing the detection on specialized hardware, we run the detection
neural network sequentially on image regions and only pool the bounding boxes,
which allows the algorithm to run efficiently on both GPU and CPU.

%% file: methodology.tex
\section{Methodology}

\subsection{Notation}

A $\bm{\mathrm{Video}}$ is a sequence   of $F$ image frames at frame number $f$. We write
\begin{equation}
    \bm{\mathrm{Video}} = \left( \bm{i}_{f} \right)_{0 \leq f < F}\mathrm{.} \label{eq:1}
\end{equation}
The $\bm{\mathrm{Video}}$ is to be annotated with $\bm{\mathrm{Annotations}}$
\begin{equation}
    \bm{\mathrm{Annotations}} = \left( \bm{\hat{a}}_{f} \right)_{0 \leq f < F} \mathrm{.} \label{eq:2}
\end{equation}
Each image frame $i_f$ is an array of pixels with width $X$, height $Y$ holding a 3
dimensional pixel vector $\bm{p}=\left[p_R~p_G~p_B\right]^{\top}$ with pixel
brightness values for Red, Green and Blue (RGB)
\begin{equation}
    \bm{i}_{f} = \left( \bm{p}_{x,y} \right)_{0 \leq x < X, 0 \leq y < Y}\label{eq:3}
\end{equation}
where $\left[ x=0~y=0 \right]^{\top}$ is the upper left corner.

Each annotation vector $\bm{\hat{a}}_f$ at frame $f$ consists of $A_f$ annotations
\begin{equation}
    \bm{\hat{a}}_{f} = \left( \bm{a}_{a,f} \right)_{0 \leq a < A_f} 
    \mathrm{.}
    \label{eq:4}
\end{equation}

Each annotation $\bm{a}_{a,f}$ includes a label $\bm{l}_{a}$ and location vector $\bm{b}_{a,f}$
\begin{equation}
    \bm{a}_{a,f} = \left\{  \bm{l}_a, \bm{b}_{a,f} \right\}\mathrm{.}\label{eq:5}
\end{equation}

While \emph{labelme} \cite{labelme2016} supports polygonal annotations for semantic
segmentation, both \emph{Smart-labelme} \cite{smart-labelme2020} and our solution
track instances by their bounding box. Polygonal annotations are scaled in $x$
and $y$ direction based on the bounding box size and shape. Therefore, each $\bm{b}$
is a vector with
\begin{equation}
    \bm{b} = \left[ x_b~y_b~X_b~Y_b \right]^{\top}\label{eq:6}
\end{equation}
\noindent where $\left[x_b~y_b\right]^{\top}$ encodes the upper left corner
and $\left[X_b~Y_b\right]^{\top}$ encodes the width and height of the
rectangular bounding box in pixels.

The label $\bm{l}$ is a textual representation, which can be used to store
information about the annotated object instance, such as its type, identity,
etc.

A crop $\bm{C}\left(\right)$ is a sub-image that is a part of a larger image $\bm{i}$.  We define $\bm{C}\left(\right)$
\begin{equation}
    \bm{C}\left( \bm{i},\bm{b}\right) = \left( \bm{\hat{c}}_{x_c,y_c} \right) = \left( \bm{p}_{\left(x_{c}+x_b\right),\left(y_{c}+y_b\right)} \right)_{0 \leq x_c < X_b, 0 \leq y_c < Y_b} 
\label{eq:7}
\end{equation}
as the tensor that maps the submatrix of a frame $\bm{i}$ defined by the
bounding box coordinates of $\bm{b}$. While $\bm{C}\left(\right)$ is only
defined within the dimensions of $\bm{i}$, we define $\bm{C_0}\left(\right)$
such that pixels outside of the dimensions of $\bm{i}$ are simply considered
black ($p_R=p_G=p_B=0$).

Similar, a scale $\bm{S}\left(\right)$ is an image similar to another image
$\bm{i}$, which has its resolution changed to $\left[ X_s~ Y_s\right]^{\top}$.
This can be defined by a remapping of pixel coordinates with tensor
$\bm{S}\left(\right)$
\begin{equation}
    \bm{S}\left( \bm{i}, \left[ X~Y \right]^{\top}, \left[ X_s~ Y_s\right]^{\top}\right)
    = \left( \bm{\hat{s}}_{x_s,y_s} \right) 
    = \left( \bm{p}_{\left(\left\lfloor{} x_s \frac{X}{X_s}\right\rfloor{}\right),\left(\left\lfloor{} y_s \frac{Y}{Y_s}\right\rfloor{}\right)} \right)_{0 \leq x_s < X_s, 0 \leq y_s < Y_s}\mathrm{,}
\label{eq:8}
\end{equation}
where $\lfloor~\rfloor$ rounds to the smaller integer.  For both tensors, there
are matching functions $c\left(\right),s\left(\right)$ and their inverse
$c^{-1}\left(\right),s^{-1}\left(\right)$ to map pixel coordinates
$\left[x~y\right]^{\top}$ or a bounding box
$\bm{b}_0=\left[x_{b0}~y_{b0}~X_{b0}~Y_{b0}\right]^{\top}$ between the image
and the crop/scale with
\begin{equation}
    \left[ x_c~y_c\right]^{\top} = c\left(\left[x~y\right]^{\top},\bm{b}\right) = \left[\left(x-x_b\right)~\left(y-y_b\right)\right]^{\top}
    \mathrm{,}\label{eq:9}
\end{equation}
\begin{equation}
    \bm{b}_c = c\left(\bm{b}_0,\bm{b}\right) = \left[\left(x_{b0}-x_b\right)~\left(y_{b0}-y_b\right)~X_{b_0}~Y_{b_0}\right]^{\top}
   \mathrm{,}\label{eq:9.5}
\end{equation}
\begin{equation}
    \left[ x_s~y_s\right]^{\top} = s\left(\left[x~y\right]^{\top},\left[X~Y\right]^{\top},\left[X_s~X_s\right]^{\top}\right) = \left[\left(x \frac{X_s}{X}\right)~\left(y \frac{Y_s}{Y}\right)\right]^ {\top}
   \mathrm{,}\label{eq:10}
\end{equation}
\begin{equation}
    \bm{b}_s = s\left( \bm{b}_0,\left[X~Y\right]^{\top},\left[X_s~X_s\right]^{\top}\right) = \left[\left(x_{b0} \frac{X_s}{X}\right)~\left(y_{b0} \frac{Y_s}{Y}\right)~\left(X_{b0} \frac{X_s}{X}\right)~\left(Y_{b0} \frac{Y_s}{Y}\right) \right]^{\top}
   \mathrm{.}\label{eq:10.5}
\end{equation}
Note: In practice, for $\bm{S}\left(\right)$ we use interpolating methods
implemented in both OpenCV and PyTorch to avoid aliasing effects.

\subsection{Multi Scale Tiling Approach}

\begin{figure}
\begin{centering}
\includegraphics[width=0.5\columnwidth]{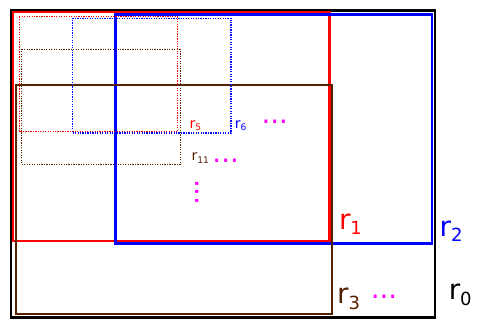}
\par\end{centering}
\caption{\label{fig:Tiling}Tiling an image into an array of overlapping tiles at different scales.}
\end{figure}

A common approach in computer vision is to divide the image $\bm{i}$ systematically
into overlapping tiles of different size. This is used among others by
\cite{10.1007/978-3-319-46448-0_2,8280135} and also the majority of Deep
Convolutional Neural Networks (CNN).

A visual operation, for example feature or object detection, can then be
executed in parallel on all these tiles, across the image at both large and
small scale. The approach is also called an ``Optical Pyramid'' and can be
considered a manifestation of the ``divide and conquer'' paradigm often found
in computer science.
If a tile $\bm{r}$ has size $\left[X_r~Y_r\right]^{\top}$, then $\bm{r_0}$
would typically represent the whole image with
\begin{equation}
    \bm{r_0} = \bm{S}\left(\bm{i},\left[ X~Y \right]^{\top},\left[ X_r~Y_r \right]^{\top} \right)\mathrm{.}
\end{equation}
All other tiles are based on a cropped tile area $\bm{b}_n=\left[x_{b_n}~y_{b_n}~X_{b_n}~Y_{b_n}\right]^{\top}$
\begin{equation}
    \bm{r_n} = \bm{S}\left(\bm{C}\left(\bm{i},\bm{b}_n \right),\left[ X_{b_n}~Y_{b_n} \right]^{\top},\left[ X_r~Y_r \right]^{\top} \right)
\end{equation}
The tiling, shown in Fig. \ref{fig:Tiling} is typically done with a recursive
algorithm that calculates all $\bm{b}_n$ for different sizes to cover the whole
image space. This can be either overlapping or non-overlapping depending on the
use case. Typically every layer of tiles has approximately $4$ times the tiles
as the previous at half the tile dimensions, while the number of layers depends
on the resolution of the original image and the minimum tile size. The latter
is often identical to $\left[X_r~Y_r\right]^{\top}$ and yields an array of $R$
tile boxes $\bm{T}$ with

\begin{equation}
    \bm{T} = \left( \bm{b}_r \right)_{0 \leq r < R}  = \bm{\mathrm{Tile}}\left( \left[X~ Y\right]^{\top}, \left[X_r~Y_r\right]^{\top} \right) \mathrm{.}
\end{equation}

\subsection{Manual Annotation\label{manual}}

\emph{Labelme} \cite{labelme2016} is optimized to streamline the process for human
annotators. One video frame $\bm{i}_f$ is always displayed prominently on the
screen. The user can navigate between frames with the arrow keys on the
keyboard and in the spacial dimension with keyboard + scroll wheel, which
allows very fast navigation. Ctrl + scroll wheel can zoom in or out to magnify
areas in high resolution images, while the Alt key or no key are used to pan
horizontally or vertically. Navigating to a specific $f$ can be done fast by
mouse-click on a timeline which highlights all frames already annotated. The
user can add annotations by entering the appropriate mode (Ctrl+R) and then
clicking diagonally opposite corners of a rectangle. A prompt box will appear
in which the user can select or type a label for the new annotation. The
previous annotation is pre-selected which is very fast if many identical labels
are given. However, modification of the label is necessary if every instance is
supposed to receive distinct identifiers. For the latter case, an experienced
and well rested annotator can achieve a speed of under 5 seconds per
annotation. For identical labels, a speed of under 4 seconds is possible. These
values are approximate, evaluated in self-test and can increase drastically
depending on the visual difficulty of the individual annotation task.

\subsection{Assisted Tracking\label{assisted}}

\emph{Smart-labelme} \cite{smart-labelme2020} adds the ability to track objects
between subsequent frames using a method for parametric image alignment
\cite{4515873}, which is implemented in the OpenCV library \cite{6240859}.
Using this to track objects works, if these are visually distinctive and do not
change appearance drastically. The method is not well suited to track organic
structures that undergo morphological changes such as a bird flapping its wings
or a horse moving its legs while walking. The process involves selecting one or
more existing annotations and then clicking ``track-polygon''. When the user then
navigates to a different video frame which does not yet have an annotation with
the same labels, the tracking algorithm is applied to find the corresponding
bounding box in the new frame and creating new object annotations. Even if the
tracking is imperfect, the human annotator only needs to adjust bounding box
corners for existing annotations, with no need to type. The algorithm
runs sequentially on CPU and is comparably slow. The annotations in the first
video frame $\bm{\hat{a}}_0$ need to be made manually as described in Sec.
\ref{manual}.

\subsection{Assisted Tracking with Re$^3$}

Re$^3$ \cite{8253805} is a recurrent siamese network, utilizing
Long-term-Short-term memory (LSTM) nodes to remember the appearance of tracked
object instances, while simultaneously observing two cropped images.
$\bm{C}\left( i_f, \bm{\tilde{b}}_{a,f}\right),\bm{C}\left( i_{f+1},
\bm{\tilde{b}}_{a,f}\right)$. Since Re$^3$ has internal memory, for each
annotation $\bm{a}_{a,f}$ we also need to store internal memory $\bm{R}_{a,f}$,
which will be updated by the tracker in every tracking transition. We can
represent Re$^3$ as a function, based on a crop region $\bm{\tilde{b}}_{a,f}$
as
\begin{equation}
\left\{ \bm{\hat{b}}_{a,f+1} , \bm{R}_{a,f+1} \right\} = 
\bm{\mathrm{Re}^3}\left(
    \bm{C_0}\left(\bm{i}_{f},\bm{\tilde{b}}_{a,f}\right),\bm{C_0}\left(\bm{i}_{f+1},\bm{\tilde{b}}_{a,f}\right),\bm{R}_{a,f}
\right)\mathrm{.}
\label{eq:11}
\end{equation}
Re$^3$ provides the new coordinates $\bm{\hat{b}}_{a,f+1}$ for the tracked objects in
the new cropped region corresponding to a object that occupies the exact center
in the old cropped region. For any annotation
$\bm{a}_{a,f}=\left\{\bm{l}_a,\bm{b}_{a,f}\right\}$ and $\bm{b}_{a,f}=\left[
    x_b~y_b~X_b~Y_b\right]^{\top}$ the crop region is always twice as large as
the to be tracked object instance, with
\begin{equation}
    \bm{\tilde{b}}_{a,f} = \left[ x_{\tilde{b}}~y_{\tilde{b}}~X_{\tilde{b}}~Y_{\tilde{b}} \right]
    =\left[ 
        \left( x_b-\left\lfloor\frac{X_b}{2}\right\rfloor \right)~\left(y_b-\left\lfloor\frac{Y_b}{2}\right\rfloor \right)~
		2X_b~2Y_b ]
    \right]^{\top}\mathrm{.}
	\label{eq:12}
\end{equation}
Which leads to annotation $\bm{a}_{a,f+1}$ based on $ \bm{\hat{b}}_{a,f+1}=\left[x_{\hat{b}}~y_{\hat{b}}~X_{\hat{b}}~Y_{\hat{b}}\right]^{\top}$ and $\bm{\tilde{b}}$
\begin{equation}
	\bm{a}_{a,f+1}=\left\{ 
        \left[ \left(x_{\hat{b}}+x_{\tilde{b}}\right)~ \left(y_{\hat{b}}+y_{\tilde{b}}\right)~X_{\hat{b}}~Y_{\hat{b}} \right]^{\top},\bm{l}_a
	\right\}\mathrm{.}
	\label{eq:13}
\end{equation}
If $\bm{R}_{a,f}$ is not known, we first initialize the tracker with
\begin{equation}
\left\{ \ldots , \bm{R}_{a,f} \right\} = 
\bm{\mathrm{Re}^3}\left(
    \bm{C_0}\left(\bm{i}_{f},\bm{\tilde{b}}_{a,f}\right),\bm{C_0}\left(\bm{i}_{f},\bm{\tilde{b}}_{a,f}\right),\bm{0}
\right)\mathrm{.}
\label{eq:14}
\end{equation}

This allows tracking equivalent to \emph{Smart-labelme} but using a modern, fast,
recurrent deep convolutional method. Re$^3$ suffers from drift, but drift only
manifests itself after some time, at which the user might have to adjust the
bounding box manually to correct for the error.

\subsection{Assisted Detection with SSD-Multibox\label{ssd}}

SSD-Multibox is a Deep Convolutional Neural Network (CNN) for object detection.
It operates with a fixed input size 
$\left[X_r~Y_r\right]^{\top}$ of $300\mathrm{px} \times 300\mathrm{px}$.
This input image is divided using $\bm{\mathrm{Tile}}$ into
$T_\mathrm{SSD}=8732$ overlapping receptive fields, called prior boxes. For
each such box the network outputs $\bm{o}_n$ with $\bm{b}_n=\left[
    x_{b_n}~y_{b_n}~X_{b_n}~Y_{b_n}~\right]^{\top}$ for a potential object instance
candidate, as well as $K$ confidence scores $\bm{k}_n$ used for classification
into $K$ object classes, so
\begin{equation}
    \left(\bm{o}_n\right)_{0 \leq n < T_\mathrm{SSD}} = \bm{\mathrm{SSD}}\left(\bm{i}\right) 
\end{equation}
with
\begin{equation}
    \bm{o}_n = \left\{ \bm{b}_n, \bm{k}_n \right\}\mathrm{.}\label{eq:21}
\end{equation}
Typically non-maximum suppression (NMS) is applied to only report the most
confident of any overlapping array of detections for each class. However, we defer
this step. When analyzing high resolution video, it is not practical to
downscale each frame to the network resolution. Instead we apply
$\bm{\mathrm{Tile}}$ to
divide each image frame $\bm{i}_f$ into $D$ overlapping tiles $\bm{b}_{d}=\left[ x_{b_{d}}~y_{b_{d}}~X_{b_{d}}~Y_{b_{d}}\right]^{\top}$,
then run SSD on each so
\begin{equation}
    \bm{i}_{f,d} = \bm{S}\left( \bm{C}\left( \bm{i}_f, \bm{b}_{d} \right), \left[ X_{b_{d}}~Y_{b_{d}}\right]^{\top}, \left[ X_r~Y_r\right]^ {\top} \right) \forall \bm{b}_{d} \in \bm{\mathrm{Tile}}\left( \left[ X~Y\right]^{\top}, \left[ X_r~Y_r \right]^{\top} \right)\mathrm{.}
\end{equation}
In each array of detections
$\bm{\mathrm{SSD}}\left(\bm{i}_{f,d}\right)$, we discard any detections that
are are close to the edges of $\bm{i}_{f,d}$, since these detections are likely
truncated. Since tiles are overlapping at different scales, there always exists
another tile that covers the whole object instance. We apply
$c^{-1}\left(\right)$ and $s^{-1}\left(\right)$ to transform the remaining
bounding boxes into the coordinate frame of $\bm{i}_f$ to compute
$\bm{\hat{o}}_f$ with
\begin{equation}
    \bm{\hat{o}}_f = \bigcup_{0 \leq d < D} c^{-1}\left(s^{-1}\left(\bm{\mathrm{remove\_truncated}}\left(\bm{\mathrm{SSD}}\left(\bm{i}_{f,d}\right)\right)\right)\right)\mathrm{.}
\end{equation}

\subsubsection{Non Maximum Suppression}

\begin{figure}[t]
\begin{centering}
\begin{tabular}{c}
\includegraphics[width=0.3\columnwidth]{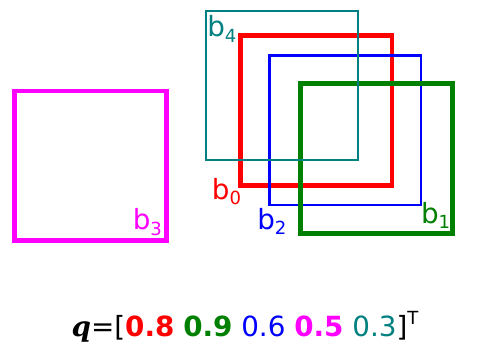}\\
\end{tabular}
\begin{tabular}{r|c|c|c|c|c|}
\multicolumn{6}{c}{$\bm{D}$:}\\
$~$ & $\bm{b}_0$ & $\bm{b}_1$ & $\bm{b}_2$ & $\bm{b}_3$ & $\bm{b}_4$\\
\hline
$\bm{b}_0$ & 1 & 0 & 1 & 0 & 1\\
\hline
$\bm{b}_1$ & 0 & 1 & 1 & 0 & 0\\
\hline
$\bm{b}_2$ & 1 & 1 & 1 & 0 & 0\\
\hline
$\bm{b}_3$ & 0 & 0 & 0 & 1 & 0\\
\hline
$\bm{b}_4$ & 1 & 0 & 0 & 0 & 1\\
\hline
\end{tabular}
\hspace{1pt}
\begin{tabular}{r|c|c|c|c|c|}
\multicolumn{6}{c}{$\bm{\hat{D}}$:}\\
$~$ & $q_0$ & $q_1$ & $q_2$ & $q_3$ & $q_4$\\
\hline
$\bm{q_0}$ & $\underline{\bm{0.8}}$ & $0$ & $0.6$ & $0$ & $0.3$\\
\hline
$\bm{q_1}$ & $0$ & $\underline{\bm{0.9}}$ & $0.6$ & $0$ & $0$\\
\hline
$q_2$ & $0.8$ & $\bm{0.9}$ & $\underline{0.6}$ & $0$ & $0$\\
\hline
$\bm{q_3}$ & $0$ & $0$ & $0$ & $\underline{\bm{0.5}}$ & $0$\\
\hline
$q_4$ & $\bm{0.8}$ & $0$ & $0$ & $0$ & $\underline{0.3}$\\
\hline
\end{tabular}
\par\end{centering}
\caption{\label{fig:NMS}Matrices $\bm{D}$ and $\bm{\hat{D}}$ for example
detections $\bm{b}$ with confidence scores
$\bm{q}=\left[0.8~0.9~0.6~0.5~0.3\right]^{\top}$. NMS is performed by selecting
only those $b_n$ for which $q_n$ (\underline{underlined})
equals the row-maximum ({\bfseries{}bold}).}
\end{figure}

Storing $\bm{\hat{o}}_f$ in a PyTorch tensor, we loop through each of $K$
object classes. For each class, we sort $\bm{\hat{o}}_f$ by confidence score
and truncate the $L=5000$ highest confidence detections. This puts an upper
bound on the computational effort.  We then have $L$ bounding boxes
$\bm{b}=\left( \bm{b}_l \right)_{0 \leq l < L}$ and $L$ confidence scores
$\bm{q}=\left( q_{l} \right)_{0 \leq l < L}$. We calculate
$\bm{J}=\left(j\right)_{m,n}$ which is an $L \times L$ matrix of the
intersection over union (IOU) of these bounding boxes. We derive
$\bm{D}$ as a boolean $L \times L$ matrix with
\begin{equation}
    \bm{D} = \left( d_{m,n} = \begin{cases} 1 \iff j_{m,n}>0.5\\ 0 \iff j_{m,n}\leq0.5\end{cases} \right)_{0 \leq m < L,0\leq n < L}\mathrm{,}
\end{equation}
from which we calculate $\bm{\hat{D}}$ by column wise multiplication with
$\bm{q}$ with
\begin{equation}
    \bm{\hat{D}}= \left(\hat{d}_{m,n} =  q_n d_{m,n} \right)_{0 \leq m < L,0\leq n < L} \mathrm{.}
\end{equation}
As illustrated in Fig. \ref{fig:NMS}. Maxima are those $\left\{ \bm{b}_l~q_l
\right\}$ for which the maximum of the $l$th row-vector of $\bm{\hat{D}}$ is
identical to $q_l$ itself, i.e. the row-maximum is on the diagonal of
$\bm{\hat{D}}$.

Crucially, this entire algorithm can be expressed in PyTorch tensor operations
and computed in a parallel on GPU with constant computational runtime. This is
substantially faster than typical iterative looping algorithms for NMS,
especially when implemented in Python.

For automatic annotation, we extend NMS by also excluding any detections that
have an IOU $\geq 0.5$ with existing $\bm{a}_{a,f} \in \bm{\hat{a}}_{f}$. We add
any remaining detections, computing a new $\bm{l}_a$ using the SSD class $k$
and a unique number computed from the system time and a counter.

\subsection{Drift-compensation with SSD-Multibox and Re$^3$\label{ours}}

For each Re$^3$ invocation, we run SSD on each $\bm{\tilde{b}}_{a,f}$. If a
detection is found (\ref{eq:21}) that has an IOU with $\bm{\hat{b}}_{a,f+1}$ as
computed by Re$^3$ (\ref{eq:11}) of $0.8$ or higher, regardless of class, then
we assume SSD found the object instance. We then correct $\bm{b}_{a,f+1}$ using
a weighed average of both SSD and Re$^3$ prediction and copy the label to
compute $\bm{\hat{a}}_{f+1}$.

%% file: experiments.tex
\section{Experiments}

\subsection{Method Variants}

\subsubsection{Baseline}
The baseline variant does not aid the annotator with
any computer vision technique. Annotations $\bm{\hat{a}}_0$ need to be
annotated by hand. In the original \emph{labelme} \cite{labelme2016}, no tracking
variant exists, and labels need to be manually copied to the next frame.
However, this is needless overhead and would not be suitable to evaluate the
impact of machine accelerated tracking. Therefore, we implemented a pseudo-tracker
that copies the current annotations to the next frame with
$\bm{\hat{a}}_{f+1}=\bm{\hat{a}}_{f}$ when engaged. This keeps the work flow
identical to the variants with a real tracker, however, the user needs to then
manually correct each bounding box in all subsequent frames, unless the object
instance has remained stationary.

\subsubsection{OpenCV}
The OpenCV variant is \emph{Smart-labelme} \cite{smart-labelme2020} as described in
Sec. \ref{assisted} using parametric image alignment to track object
instances. 

\subsubsection{Ours}
Our variant, \emph{Smarter-labelme}, uses Re$^3$ with SSD-Multibox drift
compensation as described in Sec. \ref{ours} to track, and the Auto-Annotation
function described in Sec. \ref{ssd} is available to the annotator to help
annotate $\bm{\hat{a}}_0$.

\subsection{Experiment Setup\label{exp}}

\begin{figure}[b!]
\begin{centering}
\includegraphics[width=\columnwidth]{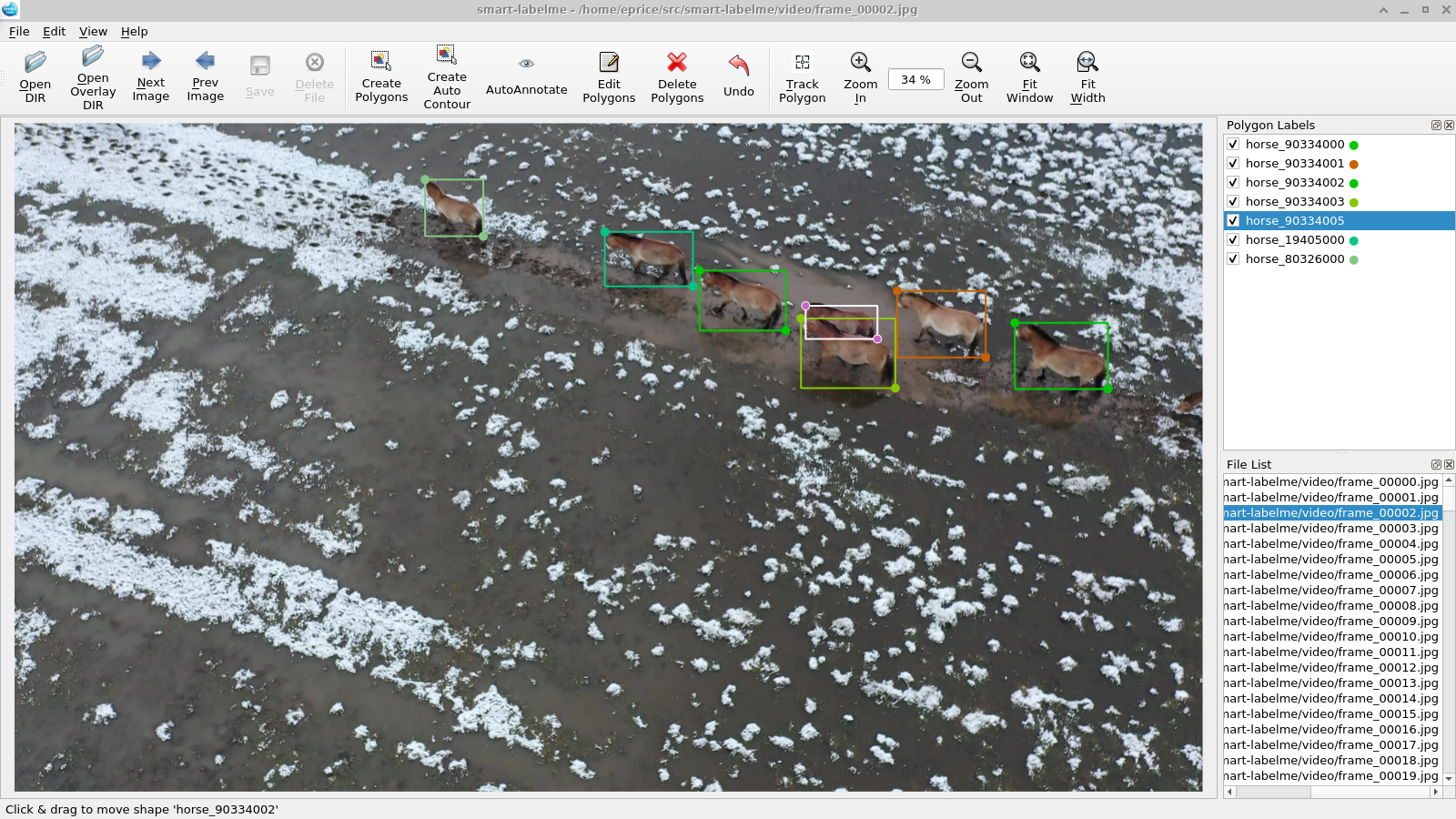}
\par\end{centering}
\caption{\label{fig:Experimentvideo}Screenshot of the experiment. Annotation of a video with walking horses.}
\end{figure}

\noindent We setup a computer with the method variants and allowed multiple
annotators to perform a video annotation task with each. For a fair comparison, 
all experiments were conducted on the same computer equipped
with a modern AMD RyZen CPU and NVIDIA GPU and a 4K monitor. The annotated
video, shown in Fig. \ref{fig:Experimentvideo} was a short 4K aerial clip of 8
horses walking on a snowy plane, shot using one of our aerial robots.  Each 
annotator was provided with an
introduction and demonstration of the labeling tool, followed by a 5 minute
practice period to familiarize themselves with the tool and the annotation
process. Then all annotations were reset and the program restarted. Each
annotator was given 10 minutes to annotate the video for each of the 3
variants (i.e., 30 min in total).  After ten minutes, the variant was switched 
and all annotations
reset. The order in which variants were evaluated was randomized to minimize
bias introduced by exhaustion or increased proficiency. The annotators were
instructed to annotate with consistent accuracy throughout the experiments. The
number of successfully annotated object instances were counted for each method
variant and annotator.

\section{Results and Discussion}

\begin{figure}[b]
\begin{centering}
\includegraphics[width=0.48\columnwidth]{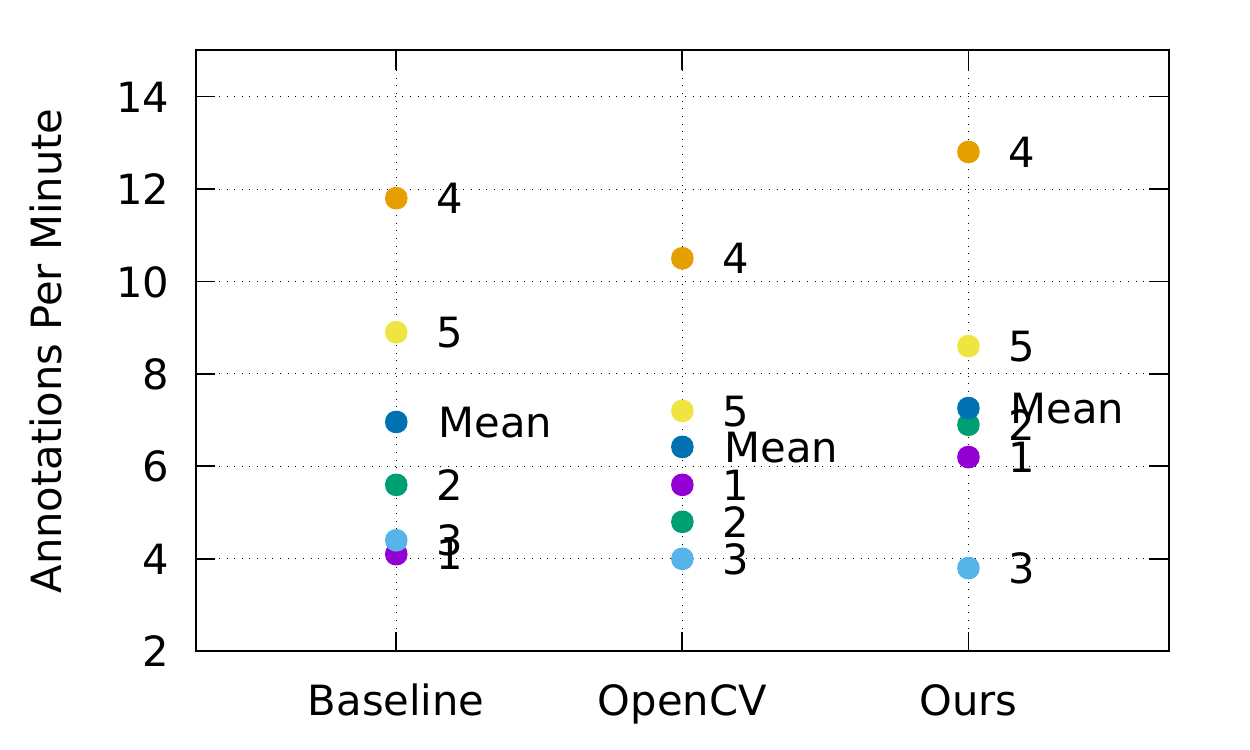}\includegraphics[width=0.48\columnwidth]{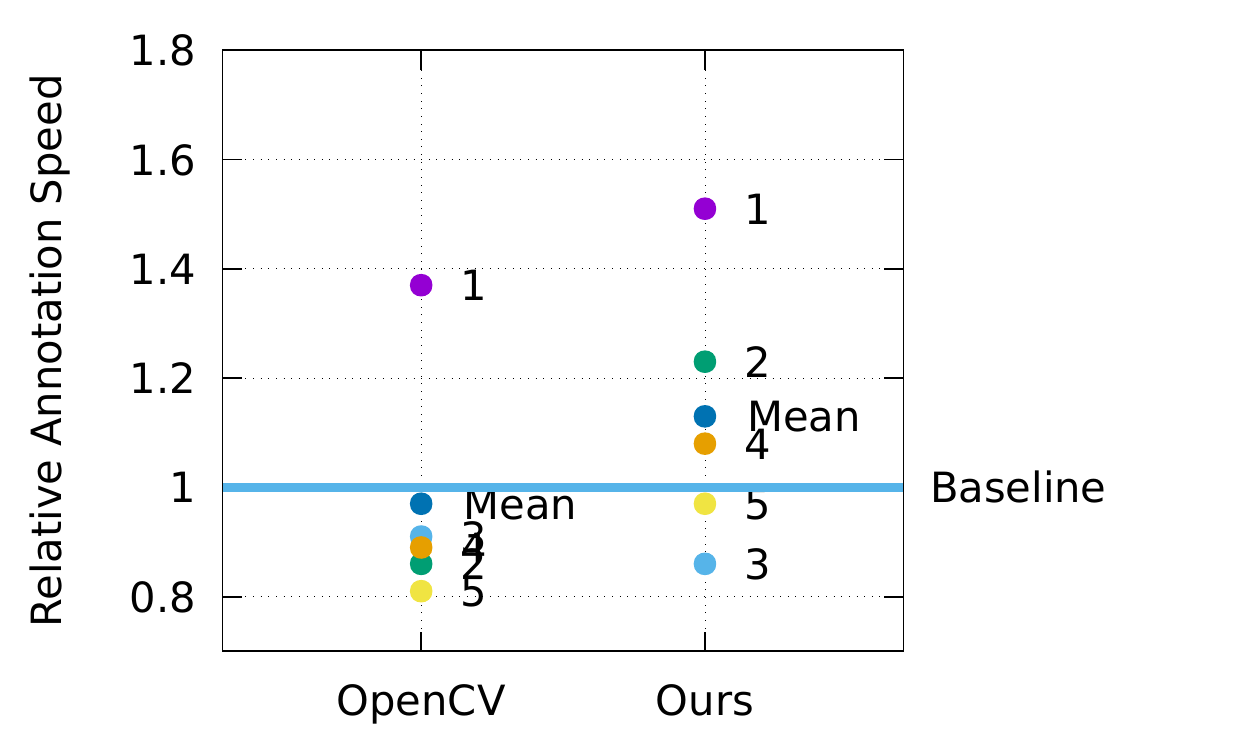}
\par\end{centering}
\caption{\label{fig:Results}Visualized experiment results. The absolute and relative annotation speed of the respective annotators are displayed, compared to baseline.}
\end{figure}

We conducted experiments as described in Sec. \ref{exp} with 5 different human
annotators. The results are shown in Table \ref{res}. On that video, the
annotators achieved on average 13\% faster annotation speeds with our method
than the baseline. \emph{Smart-labelme}
\cite{smart-labelme2020} using OpenCV was on average 3\% slower than the
baseline. As seen in Fig. \ref{fig:Results}, there were noticeable differences between the annotators. Annotator 3
reported, that they spent a lot of time re-labeling and correcting auto
annotations made with SSD-Multibox in the first frame, which had a comparably
large impact due to their slow annotation speed.  All annotators did manual
corrections to the trackers labels in every frame as needed, but consistently
reported that this process took longer if the tracking error was larger. This
explains why the baseline method, which simply copies the previous frames
labels to the next outperformed OpenCV, which had a tendency to occasionally
make large tracking errors. Another contributing factor was the long processing
time of OpenCV which added several seconds to every annotated frame. The deep
methods tracked significantly faster since they used GPU and were more
accurate, while the baseline did not spend any time processing between frames.

\begin{table}[t]
\caption{\label{res}Results of the annotation experiments.}
\begin{center}\scalebox{0.75}{
\begin{tabular}{cccccc}
\hline
Annotator&Seq.&Exp.&Frames&Annotations&Speedup\\
\hline
&1&Baseline&6&41&1.0\\
1&2&\bfseries{Ours}&\bfseries{8}&\bfseries{62}&\bfseries{1.51}\\
&3&OpenCV&7&56&1.37\\
\hline
&1&OpenCV&6&48&0.86\\
2&2&\bfseries{Ours}&\bfseries{8}&\bfseries{69}&\bfseries{1.23}\\
&3&Baseline&7&56&1.0\\
\hline
&1&\bfseries{Baseline}&\bfseries{6}&\bfseries{44}&\bfseries{1.0}\\
3&2&OpenCV&5&40&0.91\\
&3&Ours&5&38&0.86\\
\hline
&1&\bfseries{Ours}&\bfseries{16}&\bfseries{128}&\bfseries{1.08}\\
4&2&Baseline&15&118&1.0\\
&3&OpenCV&14&105&0.89\\
\hline
&1&OpenCV&9&72&0.81\\
5&2&\bfseries{Baseline}&\bfseries{12}&\bfseries{89}&\bfseries{1.0}\\
&3&Ours&11&86&0.97\\
\hline
\hline
\multicolumn{2}{c}{}&OpenCV&8.2&64.2&0.97\\
\multicolumn{2}{c}{Mean}&Baseline&9.2&69.6&1.0\\
\multicolumn{2}{c}{}&\bfseries{Ours}&\bfseries{9.6}&\bfseries{72.6}&\bfseries
{1.13}\\
\hline
\end{tabular}}
\end{center}
\end{table}

The experiments show that using our proposed approach, we can speed up
the annotation process, while keeping the processing overhead low. Secondly, 
humans are faster with a trivial method (copying bounding boxes and manually 
adjusting them) than correcting a bad automatic label. During tests we also 
found cases where, in a complex scene, a large number of objects were tracked 
correctly, while some objects due to their visual appearance, upset the tracker 
and resulted in consistent bad labels. The annotation tool should, therefore, 
allow the annotator to disable the accelerated tracker for individual objects 
if they notice such a malfunction, and revert to the trivial method of copying 
bounding boxes between frames. We have therefore added this functionality to 
our open source implementation.

The benefit of our approach depends greatly on the required annotation quality. 
In our experiment, the annotators aspired to make pixel exact annotations which 
required careful monitoring and corrections to every bounding box in every 
frame. If rough annotations are sufficient, both the validation process and the 
correction - if needed - can be significantly faster. The performance of 
detector and tracker might often be sufficient for the task. In this case, the 
processing time of the tracker becomes a major bottleneck for annotation and 
additional effort is required to optimize the software for speed. On the other 
hand, if very high quality annotations are needed, the overall speed can increase 
if the quality of machine accelerated annotations is improved. This can be done by 
fine-tuning the detector on already annotated
data.

%% file: conclusion.tex
\section{Conclusion and Outlook}

We presented an accelerated video annotation method for instance labeling in
videos. Our method combines deep networks for detection and tracking to reduce 
drift in annotation over frames, thereby substantially increasing annotation 
speed. We validated this through experiments where we compared our 
approach with baseline methods, and discussed how to maximize utility depending
on the required accuracy in annotations. We provided open-source code of our
method for the benefit of the community. Future work may involve online
training of the detector on the objects being tracked in real time, using the
already labeled annotations as training data.

%% file: bibliography.tex
 \bibliographystyle{IEEEtran}
\phantomsection\addcontentsline{toc}{section}{\refname}\bibliography{paper}